# Using ROBDDs for Inference in Bayesian Networks with Troubleshooting as an Example

**Thomas D. Nielsen**   **Pierre-Henri Wuillemin**   **Finn V. Jensen**   **Uffe Kjærulff**

Department of Computer Science
Aalborg University
Fredrik Bajers Vej 7C, DK-9220 Aalborg Ø, Denmark

## Abstract

When using Bayesian networks for modelling the behavior of man-made machinery, it usually happens that a large part of the model is deterministic. For such Bayesian networks the deterministic part of the model can be represented as a Boolean function, and a central part of belief updating reduces to the task of calculating the number of satisfying configurations in a Boolean function. In this paper we explore how advances in the calculation of Boolean functions can be adopted for belief updating, in particular within the context of troubleshooting. We present experimental results indicating a substantial speed-up compared to traditional junction tree propagation.

## 1 INTRODUCTION

When building a Bayesian network model it frequently happens that a large part of the model is deterministic. This happens particularly when modelling the behavior of man-made machinery. Then the situation is that we have a deterministic kernel with surrounding chance variables, and it seems excessive to use standard junction tree algorithms for belief updating. First of all, the calculations in the deterministic kernel are integer calculations and double precision calculations are unnecessary complex. However, there may be room for further improvements. If the deterministic part of the model is represented as a Boolean function, we may exploit contemporary advances in calculation of Boolean functions.

A major advance in Boolean calculation is Binary Decision Diagrams, particularly Reduced Ordered Binary Decision Diagrams, ROBDDs[Bryant, 1986]. An ROBDD is a DAG representation of a Boolean function. The representation is tailored for fast calculation of values, but the representation can also be used for fast calculation of the number of satisfying configurations given an instantiation of a subset of the variables.

To be more precise: let $B(\mathcal{X})$ be a Boolean function over the Boolean variables $\mathcal{X}$, and let $\mathcal{Y} \subseteq \mathcal{X}$ with $\mathcal{Z} = \mathcal{X}\backslash\mathcal{Y}$. Define $\text{Card}_B(\bar{y})$ on a configuration $\bar{y}$ of $\mathcal{Y}$ as the number of configurations $\bar{z}$ over $\mathcal{Z}$ such that $B(\bar{y}, \bar{z}) = \text{true}(1)$. It turns out that given $\bar{y}$ an ROBDD representation of B can be constructed such that $\text{Card}_B$ can be calculated in time linear in the number of nodes in the ROBDD. However, the number of nodes in an ROBDD may be exponential in the number of variables in the domain of the Boolean function.

In this paper we exploit the ROBDD representation for propagation through a Boolean kernel in a Bayesian network, and we illustrate that a central part of this propagation is to calculate $\text{Card}_B(\bar{y})$. We use the technique on models for troubleshooting. These models are particularly well suited for ROBDD calculation as the size of the ROBDD is quadratic in the size of the domain.

In section 2 we illustrate the use of $\text{Card}_B$ for probability updating in Bayesian networks. Section 3 is a brief introduction to ROBDDs and in section 4 we show how to calculate $\text{Card}_B$ in an ROBDD. Section 5 introduces the troubleshooting task and the type of Bayesian network models used. In section 6 the deterministic kernel of these models is represented as an ROBDD and it is shown that the size of this representation is quadratic in the number of Boolean variables. In section 7 we outline the propagation algorithms for various troubleshooting tasks, and in section 8 we report on empirical results indicating a substantial speed up compared to traditional junction tree propagation.

## 2 TWO MOTIVATING EXAMPLES

To illustrate the special considerations in connection with Boolean kernels we shall treat a couple of exam-ples. First consider the situation in Figure 1.

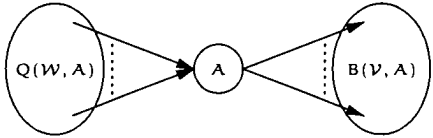

Figure 1: The Boolean variable A has a parent network of proper chance variables and a child network representing a Boolean function B.

For the situation in Figure 1 we have ($\mathcal{U} = \mathcal{W} \cup \mathcal{V} \cup \{A\}$):

$$P(\mathcal{U}) = Q(\mathcal{W}, A) \mu B(\mathcal{V}, A),$$

where $\mu = 1/\sum_{\mathcal{V} \cup \{A\}} B(\mathcal{V}, A)$ is a normalization constant. Assume we have evidence $\bar{e} = \bar{e}_\mathcal{W} \cup \bar{e}_\mathcal{V}$, where $\bar{e}_\mathcal{V}$ is a configuration $\bar{y}$ of the variables $\mathcal{Y} \subseteq \mathcal{V}$, then:

$$\begin{aligned} P(A, \bar{e}) &= \mu \sum_\mathcal{W} Q(\mathcal{W}, \bar{e}_\mathcal{W}, A) \sum_\mathcal{Z} B(\mathcal{Z}, \bar{y}, A) \\ &= \mu \sum_\mathcal{W} Q(\mathcal{W}, \bar{e}_\mathcal{W}, A) \text{Card}_B(\bar{y}, A), \end{aligned}$$

As the example illustrates, an efficient procedure for calculating $\text{Card}_B$ is central for probability updating.

If we extend the example s.t. a Boolean variable $C \in \mathcal{V}$ has a child network $R(\mathcal{T}, C)$ of proper chance variables, we get (the normalization constant is omitted):

$$P(\mathcal{U}) = Q(\mathcal{W}, A) B(\mathcal{V}, A) R(\mathcal{T}, C)$$

Assume we have evidence $\bar{e} = \bar{e}_\mathcal{W} \cup \bar{e}_\mathcal{V} \cup \bar{e}_\mathcal{T}$, where $\bar{e}_\mathcal{V}$ is a configuration $\bar{y}$ of the variables $\mathcal{Y} \subseteq \mathcal{V}$. If $\bar{e}_\mathcal{T}$ is empty then R does not contribute, and the calculations are as for Figure 1. If not, we have:

$$P(A, \bar{e}) = \sum_\mathcal{W} Q(\mathcal{W}, \bar{e}_\mathcal{W}, A) \cdot$$

$$\sum_C \left( \sum_\mathcal{Z} B(\mathcal{Z}, \bar{y}, A, C) \sum_\mathcal{T} R(\mathcal{T}, \bar{e}_\mathcal{T}, C) \right)$$

$$= \sum_\mathcal{W} Q(\mathcal{W}, \bar{e}_\mathcal{W}, A) \cdot$$

$$\left( \sum_\mathcal{Z} B(\mathcal{Z}, \bar{y}, A, C = y) \sum_\mathcal{T} R(\mathcal{T}, \bar{e}_\mathcal{T}, C = y) \right.$$

$$\left. + \sum_\mathcal{Z} B(\mathcal{Z}, \bar{y}, A, C = n) \sum_\mathcal{T} R(\mathcal{T}, \bar{e}_\mathcal{T}, C = n) \right)$$

$$= \sum_\mathcal{W} Q(\mathcal{W}, \bar{e}_\mathcal{W}, A) \cdot$$

$$\left( \text{Card}_B(\bar{y}, A, C = y) \sum_\mathcal{T} R(\mathcal{T}, \bar{e}_\mathcal{T}, C = y) \right.$$

$$\left. + \text{Card}_B(\bar{y}, A, C = n) \sum_\mathcal{T} R(\mathcal{T}, \bar{e}_\mathcal{T}, C = n) \right)$$

Again, calculation of $\text{Card}_B$ is part of belief updating.

## 3  BOOLEAN FUNCTIONS AND ROBDDS

This section is a survey of classical logic in the context of binary decision diagrams.

The classical calculus for dealing with truth assignments consists of *Boolean variables*, the constants *true* (1) and *false* (0) and the operators $\wedge$ (*conjunction*), $\vee$ (*disjunction*), $\neg$ (*negation*), $\Rightarrow$ (*implication*) and $\Leftrightarrow$ (*bi-implication*). A combination of these entities form a Boolean function and the set of all Boolean functions is known as *propositional logic.*

A truth assignment to a Boolean function B is the same as fixing a set of variables in the domain of B, i.e., if X is a Boolean variable in the domain of B, then X can be assigned either 0 or 1 (denoted $[X \mapsto 0]$ and $[X \mapsto 1]$, respectively).

A Boolean function is said to be a *tautology* if it yields true for all truth assignments, and it is *satisfiable* if it yields true for at least one truth assignment.

Let $X \to Y_1, Y_2$ denote the *if-then-else* operator. Then $X \to Y_1, Y_2$ is true if either X and $Y_1$ are true or X is false and $Y_2$ is true; the variable X is said to be the *test expression*. More formally we have:

$$(X \to Y_1, Y_2) \equiv (X \wedge Y_1) \vee (\neg X \wedge Y_2)$$

All operators in propositional logic can be expressed using only this operator and this can be done s.t. tests are only performed on unnegated variables.

**Definition 1.** An *If-then-else Normal Form (INF)* is a Boolean function built entirely from the if-then-else operator and the constants 0 and 1 s.t. all tests are performed only on variables.

Consider the Boolean function B and let $B[X \mapsto 0]$ denote the Boolean function produced by substituting 0 for X in B. The *Shannon expansion* of B w.r.t. X is defined as:

$$B \equiv X \to B[X \mapsto 1], B[X \mapsto 0]$$

From the Shannon expansion we get that any Boolean function can be expressed in INF by iteratively using the above substitution scheme on B.

By applying the Shannon expansion to a Boolean function B w.r.t. an ordering of all the variables in the domain of B we get a set of if-then-else expressions which can be represented as a *binary decision tree*. The decision tree may contain identical substructures and by "collapsing" such substructures we get a *binary decision diagram* (BDD) which is a directed acyclic graph.

The ordering of the variables, corresponding to the order in which the Shannon expansion is performed,

is encoded in the BDD hence, we say that the BDD is an *ordered binary decision diagram* (OBDD); the variables occur in the same order on all paths from the root. If all redundant tests are removed in an OBDD it is said to be *reduced* and we have a *reduced ordered binary decision diagram* (ROBDD).

**Definition 2.** A *reduced ordered binary decision diagram (ROBDD)* is a rooted, directed acyclic graph with

- one or two terminal nodes labeled 0 and 1 respectively.

- a set of non-terminal nodes of out-degree two with one outgoing arc labeled 0 and the other 1.

- a variable name attached to each non-terminal node s.t. on all paths from the root to the terminal nodes the variables respect a given linear ordering.

- no two nodes have isomorphic subgraphs.

We will use $\mathcal{E}_0$ to denote the set of 0-arcs (drawn as dashed arcs) and $\mathcal{E}_1$ to denote the set of 1-arcs (drawn as solid arcs).

**Theorem 1 ([Bryant, 1986]).** For any Boolean function $f : \{0,1\}^n \to \{0,1\}$ there is exactly one ROBDD B with variables $X_1 < X_2 < \cdots < X_n$ s.t. $B[X_1 \mapsto b_1, X_2 \mapsto b_2, \ldots, X_n \mapsto b_n] = f(b_1, b_2, \ldots, b_n), \forall (b_1, b_2, \ldots, b_n) \in \{0,1\}^n$.

From Theorem 1 we have that in order to calculate the number of satisfying configurations in a Boolean function B we can produce an ROBDD equivalent to B and then count in this structure.

In the remainder of this paper we assume that an ROBDD has exactly one terminal node labeled 1, as we are only interested in the number of satisfying configurations; in this situation we allow non-terminal nodes with out-degree one. Additionally, we will use the term "nodes" in the context of ROBDDs and "variables" when referring to a Boolean function or a Bayesian network(BN); nodes and variables will be denoted with lower case letters and upper case letters, respectively (the nodes representing a variable $X_i$ will each be denoted $x_i$ if this does not cause any confusion).

## 4 CALCULATION OF CARD$_B$ USING ROBDDS

Given an ROBDD representation of a Boolean function B, the number of satisfying configurations can be calculated in time linear in the number of nodes in the ROBDD. The algorithm basically propagates a number ($2^n$, where $n$ is the number of distinct variables in the corresponding Boolean function) from the root of the ROBDD to the terminal node. The value sent from a node (including the root) to one of its children is the value associated with that node divided by 2. The value associated with a node (except the root) is the sum of the values sent from its parents (see Figure 2).

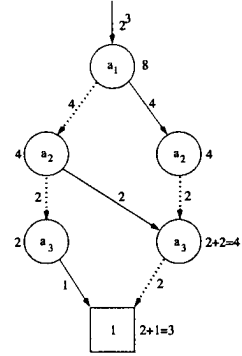

Figure 2: There are 3 satisfying configurations for the Boolean function "Exactly one variable among $A_1, A_2, A_3$ is true" represented by this ROBDD.

**Definition 3.** Let $B = (\mathcal{U}, \mathcal{E})$ be an ROBDD. Propagation in B is the computation of $v(u)$, where $u \in \mathcal{U}$ and $v : \mathcal{U} \to \mathbb{R}$ is defined as:

- $v(r) = 2^n$, where $r$ is the root in B and $n$ is the number of distinct variables in B.

- $\forall u \in \mathcal{U} \backslash \{r\} : v(u) = \frac{\sum_{p \in \Pi_u} v(p)}{2}$, where $\Pi_u$ represents the set of parents for $u$ in B.

So, in order to determine Card$_B$ for some Boolean function $B(\mathcal{U})$ we only need one propagation in the corresponding ROBDD since Card$_B = v(1)$. In case evidence $\bar{y}$ has been received on the variables $\mathcal{Y} \subseteq \mathcal{U}$ we simply modify the algorithm s.t. configurations, inconsistent with $\bar{y}$, does not contribute to the propagation, i.e., given a configuration $\bar{y}$ the function $v(u)_{\bar{y}}$ is defined as:

$$\forall u \in \mathcal{U} \backslash \{r\} : v(u)_{\bar{y}} = \frac{\sum_{p \in \Pi_u^{\bar{y}}} v(p)_{\bar{y}}}{2},$$

where $\Pi_u^{\bar{y}} = \{p \in \Pi_u | [p \notin \mathcal{Y}] \text{ or } [\bar{y}(p) = i \text{ and } (p, u) \in \mathcal{E}_i]\}$; $\bar{y}(p)$ is the state of $p \in \mathcal{Y}$ under $\bar{y}$ and $v(r) = 2^n$, $n$ being the number of distinct variables in B including those on which evidence has been received. In particular we have that Card$_B(\bar{y}) = v(1)_{\bar{y}}$. Notice, that the structure of the ROBDD is not changed when evidence is received.

The size of the ROBDD has a significant impact on the performance of the algorithm and the problem of identifying a minimal sized ROBDD is NP-complete. Thus, in the remainder of this paper we shall mainly focus on troubleshooting models as it turns out that the structure of such a model ensures that the size of the corresponding ROBDD is at most quadratic in the size of the domain.

## 5 TROUBLESHOOTING

When troubleshooting a device which is not working properly we wish to determine the cause of the problem or find an action sequence repairing the device. At any time during the process there may be numerous different operations that can be performed e.g. a component can be repaired/replaced or the status of a component can be investigated. Because such operations can be expensive and may not result in a functioning device, it is expedient to determine a sequence of operations that minimizes the expected cost and (eventually) repairs the device.

[Breese and Heckerman, 1996] presents a method to myopicly determine such a sequence. The method assumes a BN representing the device in question, and the BN is assumed to satisfy the following properties.

1. There is only one *problem defining variable* in the BN and this variable represents the functional status of the device.

2. The device is initially faulty.

3. Exactly one component is malfunctioning causing the device to be faulty (*single fault*).

A central task of troubleshooting, within the framework of [Breese and Heckerman, 1996], is the calculation of $p_i = P(C_i = \text{faulty}|\bar{e})$ which denotes the probability that component $C_i$ is the cause of the problem given evidence $\bar{e}$. So we are looking for a way to exploit the logical structure of the model when calculating the probabilities $p_i$. As such a scheme is strongly dependent on the structure of the troubleshooting model we give a syntactical definition of this concept. The definition is based on BNs: a BN consists of a directed acyclic graph $G = (\mathcal{U}, \mathcal{E})$ and a joint probability distribution $P(\mathcal{U})$, where $\mathcal{U}$ is a set of variables and $\mathcal{E}$ is a set of edges connecting the variables in $\mathcal{U}$; we use $\text{sp}(X)$ to denote the state space for a variable $X \in \mathcal{U}$. The joint probability distribution $P(\mathcal{U})$ factorizes over $\mathcal{U}$ s.t.:

$$P(\mathcal{U}) = \prod_{X \in \mathcal{U}} P(X|\Pi_X),$$

where $\Pi_X$ is the parents of $X$ in $G$. The set of conditional probability distributions factorizing $P(\mathcal{U})$ according to $G$ is denoted $\mathcal{P}$.

**Definition 4.** A troubleshooting model is a connected BN $\text{TS} = ((\mathcal{U} = \mathcal{U}_S \cup \mathcal{U}_C \cup \mathcal{U}_A, \mathcal{E}), \mathcal{P})$, where:

- The set $\mathcal{U}_S$ contains a distinct variable S with no successors, and for each $S' \in \mathcal{U}_S \setminus \{S\}$ there exists a directed path from $S'$ to $S$.

- For each variable $C \in \mathcal{U}_C$ there exists an $S' \in \mathcal{U}_S$ s.t. $C \in \Pi_{S'}$ and $\Pi_C = \emptyset$.

- For each variable $A \in \mathcal{U}_A$ there does not exist an $X \in \mathcal{U}$ s.t. $A \in \Pi_X$.

- $\text{sp}(X) = \{\text{ok}, \neg\text{ok}\}, \forall X \in \mathcal{U}_S \cup \mathcal{U}_C$.

- For each $X \in \mathcal{U}_S$: $P(x|\bar{y}) = 1$ or $P(x|\bar{y}) = 0$, $\forall x \in \text{sp}(X)$ and $\forall \bar{y} \in \text{sp}(\Pi_X)$.

The variable S is termed the problem defining variable and the variables $\mathcal{U}_S$ are termed *system variables*. The variables $\mathcal{U}_C$ (termed *cause variables*) represent the set of components which can be repaired, and the variables in $\mathcal{U}_A$ (termed *action variables*) represent user performable operations such as *observations* and *system repairing actions*; notice that $\mathcal{U}_A$ is not part of the actual system specification. In the remainder of this paper we shall extend the single fault assumption to include the system variables also. That is, if a system variable $S_i$ is faulty, then there exists exactly one variable $X \in \Pi_{S_i}$ which is faulty also (see [Skaanning et al., 1999] for further discussion of this extension and how the single fault assumption can be enforced using so-called "constraint variables").

Figure 3 depicts a troubleshooting model, where A is an action variable and S represents the problem defining variable. The variables $S_1, S_2, S_3$ and $S_4$ represent subsystems, which should be read as: the system S can be decomposed into two subsystems $S_1$ and $S_2$, and subsystem $S_1$ can be decomposed into $S_3$ and $S_4$. Component $C_1$ can cause either $S_3$ or $S_4$ to fail, whereas $C_2$ can cause either $S_2$ or $S_4$ to fail (neither $C_1$ nor $C_2$ can cause two subsystems to fail simultaneously). Notice that A is not part of the actual system model.

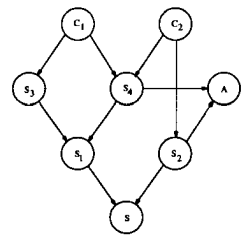

Figure 3: A troubleshooting model with five system variables, two cause variables and one action variable.

From assumption (2) and (3) we have that:

$$P(C_1 = y, \bar{e})$$
$$= P(C_1 = y, C_2 = n, \ldots, C_m = n, \bar{e})$$
$$= P(C_1 = y) \prod_{i=2}^{m} P(C_i = n)$$
$$\sum_{\mathcal{U}_S} \mu B(C_1 = y, C_2 = n, \ldots, C_m = n, \mathcal{U}_S, \bar{e})$$
$$= P(C_1 = y) \prod_{i=2}^{m} P(C_i = n) \mu \text{Card}_B(C_1 = y, \bar{e}),$$

where $B(\mathcal{U}_S, \mathcal{U}_C)$ is a Boolean function (specified in the following section) and $\mu$ is a normalization constant. Now, $P(C_1|\bar{e}) = P(C_1, \bar{e})/P(\bar{e})$ and $P(\bar{e})$ is given by:

$$P(\bar{e}) = \sum_{\mathcal{U}} \prod_{i=1}^{m} P(C_i) \mu B(C_1, \ldots, C_m, S, S_1, \ldots, S_n, \bar{e})$$

In the remainder of this paper we omit the normalization constant.

## 6 ROBDDS AS TROUBLESHOOTING MODELS

In what follows we shall assume single fault and use the truth values 1 and 0 to denote the state of a component/subsystem (1 indicates a fault).

Now, let $\Pi_{S_i}$ be the subsystems which immediately compose $S_i \in \mathcal{U}_S$ and let $\mathcal{S}_C \subseteq \mathcal{U}_S$ be the subsystems that component $C \in \mathcal{U}_C$ can cause to fail; $\mathcal{S}_C$ is the immediate successors of C. The Boolean function representing the logical kernel of a troubleshooting model $TS = ((\mathcal{U}_S \cup \mathcal{U}_C \cup \mathcal{U}_A, \mathcal{E}), \mathcal{P})$ is then given by $B(\mathcal{U}' = \mathcal{U}_S \cup \mathcal{U}_C)$:

$$F(T) = \left(T \wedge \bigotimes_{S' \in \Pi_T} S'\right) \vee \left(\neg T \wedge \bigwedge_{S' \in \Pi_T} \neg S'\right)$$
$$G(C) = C \Rightarrow \bigotimes_{T \in \mathcal{S}_C} T$$
$$M = \left(S \wedge \bigotimes_{C \in \mathcal{U}_C} C\right) \vee \left(\neg S \wedge \bigwedge_{C \in \mathcal{U}_C} \neg C\right)$$
$$B(\mathcal{U}') = \left(\bigwedge_{T \in \mathcal{U}_S} F(T)\right) \wedge \left(\bigwedge_{C \in \mathcal{U}_C} G(C)\right) \wedge M,$$

where $\bigotimes_{i=1}^{n} X_i$ denotes an *exclusive-or* between the variables $\{X_1, \ldots, X_n\}$. $F(T)$ specifies that if the system T is malfunctioning then one (and only) of its subsystems is faulty, and if the system is functioning properly then all of its subsystems are functioning properly also. $G(C)$ states that if a cause is present then one (and only one) of its subsystems ($\mathcal{S}_C$) is faulty (if a cause is not present we can not say anything about its subsystems). $M$ says that there can be either zero or at most one cause present (consistent with the system state). $B(\mathcal{U})$ is the Boolean function representing the system as a whole. Note that:

- The Boolean function is a list of expressions for local constraints and it can therefore be built in an incremental fashion.

- The Boolean function can easily be modified to represent any logical relation between the components.

- The expression ensures the single fault assumption based on the structure of the model, i.e., it is not necessary to introduce "constraint variables".

**Example 1.** The Boolean function representing the troubleshooting model depicted in Figure 3 is specified by B:

$$B_1 = ((S \wedge (S_1 \otimes S_2)) \vee (\neg S \wedge \neg S_1 \neg S_2))$$
$$B_2 = B_1 \wedge ((S_1 \wedge (S_3 \otimes S_4)) \vee (\neg S_1 \wedge \neg S_3 \neg S_4))$$
$$B_3 = B_2 \wedge (C_1 \Rightarrow (S_3 \otimes S_4))$$
$$B_4 = B_3 \wedge (C_2 \Rightarrow (S_2 \otimes S_4))$$
$$B = B_4 \wedge ((S \wedge (C_1 \otimes C_2)) \vee (\neg S \wedge \neg C_1 \wedge \neg C_2))$$

Given the ordering $S, S_1, S_2, S_3, S_4, C_1, C_2$, the ROBDD corresponding to B is depicted in Figure 4. Note that all paths from the root S to the terminal node are consistent with the ordering above.[1] □

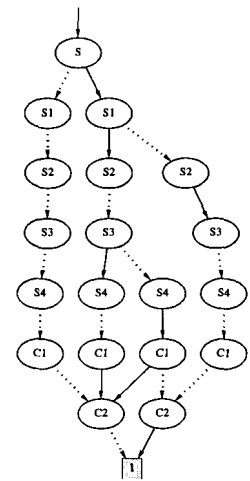

Figure 4: An ROBDD representation of the troubleshooting model depicted in Figure 3.

---

[1] The ROBDD was generated by the software tool *iben*, http://www.cs.auc.dk/~behrmann/iben/.

Now, as indicated in Section 3, the size of the ROBDD is dependent on the ordering of the variables. So we are looking for a general "rule of ordering" producing ROBDDs of "small" size.

Consider an ordering of the variables where each system variable occurs before all the variables representing its subsystems, and where all the cause variables occur last in the ordering. By constructing the ROBDD according to this ordering we get the node representing the problem defining variable as root and the nodes representing the cause variables at the bottom (see Figure 4). Moreover, we get an upper bound on the size of the ROBDD as stated in the following theorem; note that the action variables are not part of the logical kernel.

**Theorem 2.** Let $\mathsf{TS} = ((\mathcal{U} = \mathcal{U}_A \cup \mathcal{U}_S \cup \mathcal{U}_C, \mathcal{E}), \mathcal{P})$ be a troubleshooting model. Then the size of the ROBDD, representing the Boolean function $B(\mathcal{U}_S \cup \mathcal{U}_C)$, is $O(|\mathcal{U}_S|^2 + |\mathcal{U}_C|^2)$, if the ordering $\alpha : \mathcal{U}_S \cup \mathcal{U}_C \leftrightarrow |\mathcal{U}_S \cup \mathcal{U}_C|$ satisfies:

- $\forall X \in \mathcal{U}_S : \alpha(X) < \alpha(Y)$ for each $Y \in \Pi_X$.

- $\forall Z \in \mathcal{U}_C$ there does not exist an $X \in \mathcal{U}_S$ s.t. $\alpha(Z) < \alpha(X)$.

*Proof.* Assume an indexing of the layers in the ROBDD s.t. the layers containing the root node and the terminal node have index 1 and $|\mathcal{U}_S \cup \mathcal{U}_C| + 1$, respectively; a layer is the set of nodes representing a distinct variable.

Now, consider the layers consisting of system nodes but no cause nodes. The number of nodes in the i'th layer either equals the number of nodes in the i'th $-1$ layer or it has exactly one more node than the i'th $-1$ layer. This is the same as saying that at most one node in the i'th $-1$ layer branches in two; if two different nodes in a layer branched into two we would have two distinct paths from a node at a higher level to these nodes however, this contradicts the single-fault assumption due to the ordering of the nodes. Thus, the number of nodes in the layers containing system nodes is at most $\sum_{i=1}^{|\mathcal{U}_S|} i = \frac{|\mathcal{U}_S|(|\mathcal{U}_S|+1)}{2}$.

For the cause nodes, there can be at most one distinct path for each of their possible configurations. This means that the number of nodes in the layers containing cause nodes is at most $|\mathcal{U}_C|\binom{|\mathcal{U}_C|}{1} = |\mathcal{U}_C|^2$. Hence, the size of the ROBDD is $O(|\mathcal{U}_S|^2 + |\mathcal{U}_C|^2)$. □

In the ROBDDs, we have an *all-false* path from the root to the terminal node. Indeed the Boolean function is true when the model has no fault. However, we can force S to be true (faulty) to avoid this path.

## 7 PROPAGATION USING ROBDDS

For our context, we need to compute the number of satisfying configurations for each instantiation of the cause variables (see Section 5). Now, if we order the variables as described in Theorem 2 we get an ROBDD where the nodes representing the cause variables are the nodes closest to the terminal node. This means that after one propagation we can determine all the values needed, i.e., the number of configurations consistent with $C_i = y$ and evidence $\bar{e}$ is given by:

$$\mathrm{Card}_B(C_i = y, \bar{e}) = \sum_{c_i \in \mathcal{C}_i} \frac{\nu(c_i)_{\bar{e}}}{2^{\#l_i}},$$

where $\mathcal{C}_i$ is the set of nodes $c_i$ with an outgoing 1-arc and $\#l_i$ is the number of arcs on the path $l_i$ from the $c_i$ in question to the terminal node; the single-fault assumption ensures that there exist exactly one path from each $c_i$ to the terminal node which include the 1-arc emanating from $c_i$.

However, this scheme does not take user performable operations (i.e. $\mathcal{U}_A$) into account, and in the following section we extend the algorithm to include such scenarios.

### 7.1 Inserting evidence

After an action has been performed we may gain new knowledge about the system. This knowledge is incorporated into the model by instantiating the appropriate variable. If either a system variable or a cause variable is instantiated we can use the method described in Section 4. So, let $A \in \mathcal{U}_A$ be a binary variable associated with a proper conditional probability distribution $P(A|S_i)$ and assume that $A = y$ is observed. In order to take the state of A into consideration we get:

$$\begin{aligned}
&P(C_1 = y, A = y) \\
&= P(C_1 = y, C_2 = n, \ldots, C_m = n, A = y) \\
&= P(C_1 = y) \prod_{i \neq 1} P(C_i = n) \sum_{\mathcal{U}_S} (P(A = y|S_i) \\
&\quad B(C_1 = y, C_2 = n, \ldots, C_m = n, \mathcal{U}_S))
\end{aligned}$$

By expanding the sum in the above equation we get:

$$\begin{aligned}
\sum_{\mathcal{U}_S} & P(A = y|S_i) B(C_1 = y, C_2 = n, \ldots, C_m = n, \mathcal{U}_S) \\
&= P(A = y|S_i = y) \mathrm{Card}_B(C_1 = y, S_i = y) \\
&\quad + P(A = y|S_i = n) \mathrm{Card}_B(C_1 = y, S_i = n)
\end{aligned} \quad (1)$$

Thus, with one piece of evidence we can retrieve the probabilities with two propagations. However, if we have a set of actions $\mathcal{U}_A' \subseteq \mathcal{U}_A$ with parents $\mathcal{U}_S' \subseteq \mathcal{U}_S$ we need to count the number of satisfying configurations consistent with each configuration of $\mathcal{U}'_S$. So, by using the above approach, the number of times we need to count is exponential in the number of variables on which evidence has been received.

In what follows we will consider a different algorithm, where all values can be found after one propagation. Initially we assume that evidence has been received on exactly one variable, but the algorithm can easily be generalized to any number of variables.

In order to prove the soundness of the algorithm we will use the following notation. If B is an ROBDD with root r, then $L_i = \{v | r = v_1, \ldots, v_i = v \text{ is a directed path in B}\}$ is termed the i'th layer of B; the layers $L_1$ and $L_{n+1}$ contain the root node and the terminal node, respectively. So, given a Boolean function over the variables $\mathcal{U} = \{X_1, X_2, \ldots, X_n\}$ (ordered by index), the corresponding ROBDD can be specified as $B = (\mathcal{U}_B = \cup_{k=1}^{n+1} L_k, \mathcal{E} = \mathcal{E}_1 \cup \mathcal{E}_0)$; assuming that the variable $X_i$ is represented by the layer $L_i$. Now, let $f : sp(\mathcal{W}) \to \mathbb{R}$ be a function where $\mathcal{W} = \{X_i, \ldots, X_j\} \subseteq \mathcal{U}$, and assume that the variables are ordered by index. We define the following partitioning of $B = (\mathcal{U}_B = \cup_{k=1}^{n+1} L_k, \mathcal{E})$ w.r.t. f:

- The *root part* of B w.r.t. f is given by $B^r = (\mathcal{U}^r_B, \mathcal{E}^r_B)$, where $\mathcal{U}^r_B = \cup_{k=1}^{i-1} L_k$.

- The *conditioning part* of B w.r.t. f is given by $B^c = (\mathcal{U}^c_B, \mathcal{E}^c_B)$, where $\mathcal{U}^c_B = \cup_{k=i}^{j} L_k$.

- The *terminal part* of B w.r.t. f is given by $B^t = (\mathcal{U}^t_B, \mathcal{E}^t_B)$, where $\mathcal{U}^t_B = \cup_{k=j+1}^{n+1} L_k$.

For ease of exposition, we shall in the remainder of this section assume that no evidence has been received on any variable in $\mathcal{U}_S \cup \mathcal{U}_C$; the results presented can easily be generalized to this situation also.

**Algorithm 1.** *Let $B = (\mathcal{U} = \cup_{i=1}^{n+1} L_i, \mathcal{E} = \mathcal{E}_1 \cup \mathcal{E}_0)$ be an ROBDD corresponding to a Boolean function over the variables $\mathcal{U} = \{X_1, X_2, \ldots, X_n\}$, and assume that the variables are ordered by index. Let $f : sp(\mathcal{W}) \to \mathbb{R}$ be a function with $\mathcal{W} \subseteq \mathcal{U}$ and let $Q = \mathcal{W} \setminus \{X_j\}$, where $X_j \in \mathcal{W}$ is the variable with highest index.*

i) *Propagate from the root to the terminal nodes in the root part of B.*

ii) *Use the values obtained in step (i) to perform a propagation in the conditioning part of B, i.e., for each $\bar{q} \in sp(Q)$:*

  a) *Propagate to layer $L_j$.*

  b) *If there exists an arc $(p, u) \in \mathcal{E}_i$ from a node $p \in L_j$ to a node $u \in L_{j+1}$ add the value $\left(\frac{f(\bar{q}, X_j = i) v(p)_{\bar{q}}}{2}\right)$ to the value of u.*

iii) *Use the values obtained in step (ii) to propagate in $B^t$.*

Note, that the number of variables in the domain of f determines the number of iterations performed by the algorithm. In particular, if $|\mathcal{W}| = 1$ we only need one iteration.

**Theorem 3.** Let $B = (\mathcal{U} = \cup_{i=1}^{n+1} L_i, \mathcal{E} = \mathcal{E}_1 \cup \mathcal{E}_0)$ be an ROBDD and let $f : sp(\mathcal{W}) \to \mathbb{R}$ be a function where $\mathcal{W} \subseteq \mathcal{U}$. If Algorithm 1 is invoked on B, then:

$$v(1) = \sum_{\bar{w} \in sp(\mathcal{W})} f(\bar{w}) \mathrm{Card}_B(\bar{w})$$

*Proof.* Let $Q = \mathcal{W} \setminus \{X_j\}$, where $X_j \in \mathcal{W}$ is the variable with highest index. Let $\bar{q} \in sp(Q)$ and let $\Pi_u^{(\bar{q},i)} = \{p \in \Pi_u | p \notin \mathcal{W} \text{ or } (p, u) \in \mathcal{E}_i\}$. Then $\forall u \in L_{j+1}$ we have:

$$v(u) = \frac{\sum_{\bar{q} \in sp(Q)} \left(\sum_{b \in \{0,1\}, p \in \Pi_u^{(\bar{q},b)}} v(p)_{\bar{q}} f(\bar{q}, b)\right)}{2}$$

$$= \frac{\sum_{\bar{q} \in sp(Q)} \left(\sum_{b \in \{0,1\}, p \in \Pi_u^{(\bar{q},b)}} v(p)_{(\bar{q},b)} f(\bar{q}, b)\right)}{2}$$

$$= \frac{\sum_{\bar{w} \in sp(\mathcal{W})} \left(\sum_{p \in \Pi_u^{\bar{w}}} v(p)_{\bar{w}} f(\bar{w})\right)}{2}$$

$$= \frac{\sum_{\bar{w} \in sp(\mathcal{W})} f(\bar{w}) \sum_{p \in \Pi_u^{\bar{w}}} v(p)_{\bar{w}}}{2}$$

$$= \sum_{\bar{w} \in sp(\mathcal{W})} f(\bar{w}) v(u)_{\bar{w}}$$

Let $u \in L_l$, for $l > j + 1$. Suppose that $\forall p \in L_{l-1} : v(p) = \sum_{\bar{w} \in sp(\mathcal{W})} f(\bar{w}) v(p)_{\bar{w}}$. Then:

$$v(u) = \frac{\sum_{p \in \Pi_u} v(p)}{2} = \frac{\sum_{p \in \Pi_u} \sum_{\bar{w} \in sp(\mathcal{W})} f(\bar{w}) v(p)_{\bar{w}}}{2}$$

In particular we have that for $l = n + 1$:

$$v(1) = \frac{\sum_{\bar{w} \in sp(\mathcal{W})} f(\bar{w}) \sum_{p \in \Pi_u} v(p)_{\bar{w}}}{2}$$

$$= \sum_{\bar{w} \in sp(\mathcal{W})} f(\bar{w}) \mathrm{Card}_B(\bar{w})$$

Thereby completing the proof. $\square$

By performing induction in the number of operations the algorithm can easily be extended to handle multiple functions, assuming that the variables in the domain of the functions do not *overlap*; the variables in the domain of two functions f and g are said to overlap w.r.t. the ordering $\alpha$ if $\alpha(X_i) < \alpha(X_k) < \alpha(X_j)$, where $X_k$ is a variable in the domain of g, and $X_i$ and $X_j$ are the variables in the domain of f with lowest and highest index, respectively. If the variables of two functions overlap we can multiply these functions and consider the resulting function.

**Example 2.** Consider the troubleshooting model depicted in Figure 3, and assume that action $A \in \mathcal{U}_A$ is associated with the conditional probability distribution specified in Table 1; $\Pi_A = \{S_2, S_4\}$.

Table 1: The conditional probability function $P(A|S_2, S_4)$.

| $P(A = y|S_2, S_4)$ | $S_4 = 1$ | $S_4 = 0$ |
|---|---|---|
| $S_2 = 1$ | 0.3 | 0.6 |
| $S_2 = 0$ | 0.2 | 0.4 |

The ROBDD corresponding to this specification is depicted in Figure 4. In the naive approach, if $A = y$ is observed, we perform three propagations to the terminal node (one propagation for each configuration of $S_2$ and $S_4$ except for $(S_2 = 1, S_4 = 1)$ due to the single fault assumption). The resulting counts are weighted with the appropriate values and then added (see equation 1): $0 \cdot 0.3 + 4 \cdot 0.2 + 1 \cdot 0.4 + 1 \cdot 0.6 = 1.8$.

When using algorithm 1 we start off by propagating to the layer $L_3$ (the nodes representing $S_2$); after propagation, each node in $L_3$ is associated with $2^5$. We then perform two propagations to the layer $L_5$ (the nodes representing $S_4$); each propagation is conditioned on the state of $S_2$, i.e., 1 and 0, respectively. After each propagation, the resulting value is multiplied with the appropriate value from the conditional probability table and then added to the value associated with its child. So, the final value can be found with less than two full propagations (see Figure 5); note that we only perform one propagation in $B^r$ and in $B^t$. □

Step (ii) of Algorithm 1 can be optimized by starting the iteration with the variable with highest index, and then iterate in reverse order of their index. That is, when iterating over the variables $\{X_1, X_2, \ldots, X_{l-1}, X_l\}$ we can start off by propagating to the layer containing $X_l$, for some configuration of $\{X_1, X_2, \ldots, X_{l-1}\}$. The values associated with the nodes $x_{l-1}$ can then be used when propagating from the nodes $x_l$, for each instance of $x_l$. The same applies when considering variables of lower index, i.e., we can reuse previous computations. For instance, in Figure 5 we can use the value from the first iteration when computing the value $0.2 \cdot 2^2$ associated with $c_1$ (consistent with $(S_2 = 0, S_4 = 1)$).

## 8 RESULTS

We have measured the performance of the ROBDD algorithm by comparing it to the *Shafer-Shenoy algorithm* [Shafer and Shenoy, 1990] and the *Hugin algorithm* [Jensen et al., 1990] w.r.t. the number of operations performed during inference; the number of opera-tions refers to the number of additions, multiplications and divisions.

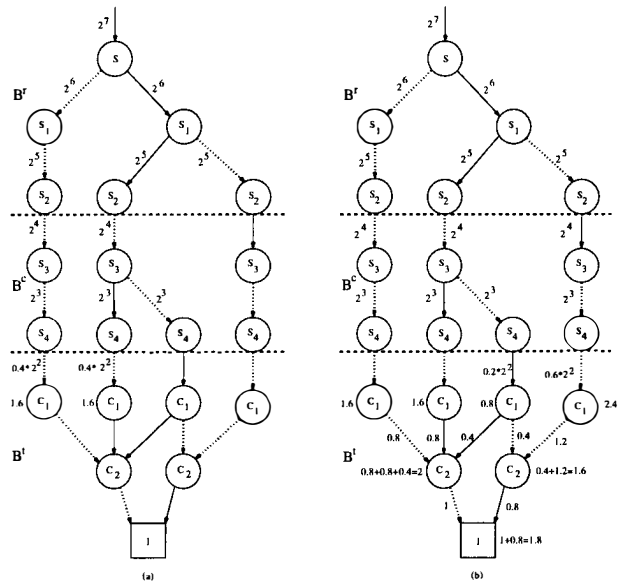

Figure 5: Figure (a) depicts the ROBDD after propagation w.r.t. the configuration $(S_2 = 0, S_4 = 0)$. Figure (b) depicts the ROBDD after the full propagation; no propagation is performed w.r.t. $(S_2 = 1, S_4 = 1)$ due to the single fault assumption.

The tests were performed on 225 randomly generated troubleshooting models (see Definition 4) which differed in the number of system variables, cause variables and action variables; the total number of variables varied from 21 to 322 and for a fixed set of variables 15 different troubleshooting models were generated. As the single fault assumption is not ensured in the troubleshooting models we augmented these models with constraint variables when using the Hugin algorithm and the Shafer-Shenoy algorithm (the single fault assumption is naturally ensured in the ROBDD architecture). Finally, evidence were inserted on the problem defining variable and on the constraint variables.

Figure 6 show plots of the number of operations performed as a function of the number of variables in the models. Note that we use a logarithmic scale on the y-axis and that the numbers on the x-axis do not represent the actual number of variables in the models. The plots show that, w.r.t. the number of operations, propagation using ROBDDs is considerably more efficient than both Shafer-Shenoy and Hugin propagation. Moreover, as indicated in Section 6, the traditional tradeoff between time and space is less apparent in the ROBDD architecture, as the space complexity is $O(|\mathcal{U}_C|^2 + |\mathcal{U}_S|^2)$.

It should be noted that the tests were designed to compare ROBDD propagation with Shafer-Shenoy and Hugin propagation, and they should not be seen as a comparison of Shafer-Shenoy propagation and Hugin propagation. In particular, we have only considered troubleshooting models and not Bayesian networks in general.

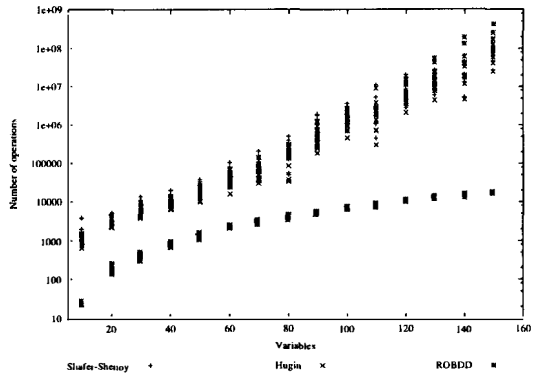

Figure 6: A plot of the number of operations performed by Hugin, Shafer-Shenoy and ROBDD propagation as a function of the number of variables in randomly generated troubleshooting models (logarithmic scale).

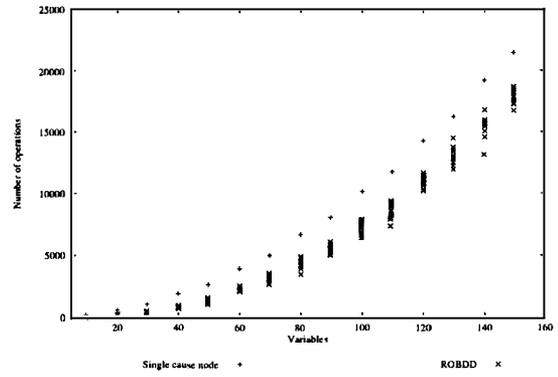

Figure 7: A plot of the number of operations performed by ROBDD propagation and Hugin propagation with a single cause node.

The efficiency of the ROBDD architecture is partly based on the single fault assumption. However, this assumption can also be exploited in certain troubleshooting models by compiling the original model $TS = ((\mathcal{U}_S \cup \mathcal{U}_C \cup \mathcal{U}_A, \mathcal{E}), \mathcal{P})$ into a secondary Bayesian network $BN = ((\mathcal{U}_A \cup \{C\} \cup \{S\}, \mathcal{E}'), \mathcal{P}')$, where C is a variable having a state for each cause variable in the original model together with a state representing the situation where no fault is present. S is a problem defining variable having C as parent, and $\mathcal{U}_A$ is the set of action variables in the original model each having C as parent. We have compared the ROBDD architecture with this approach using the randomly generated troubleshooting models from the previous tests (see Figure 7).

By using this secondary representation the speed-up is less apparent. However, if we allow multiple faults then this representation can not be used. Moreover, a troubleshooting model allowing multiple faults will in general not be simpler than a model with no constraints on the number of faults. In the case of ROBDDs, assume that the single fault assumption still applies to the system variables and consider the case where exactly $m$ components can fail simultaneously; $m$ is generally "small". In this situation the number of nodes in the layers containing system nodes does not change but the number of nodes in the layers containing cause nodes do: there can be a distinct path for each configuration of the cause nodes so the number of nodes in the layers containing cause nodes is at most $|\mathcal{U}_C|\binom{|\mathcal{U}_C|}{m}$. Hence the size of the ROBBD is $O(|\mathcal{U}_S|^2 + |\mathcal{U}_C|\binom{|\mathcal{U}_C|}{m})$; note that in an ROBDD there does not exist two nodes having isomorphic subgraphs so the size of the ROBDD is usually much smaller.

Now, as the complexity of propagation in an ROBDD is linear in its size, the maximum number of operations performed for $m = 2$ increases by a factor of $\frac{n-1}{2}$; with $m$ faults the maximum number of operations increases by a factor of $\frac{\prod_{i=n-m}^{n-1} i}{m!}$. This corresponds to adding a constant value to the ROBDD plots in Figure 6 since we use a logarithmic scale on the y-axis.

Furthermore, if we redefine the $m$-faults assumption to cover *at most* $m$ faults then the number of nodes in the layers containing cause nodes is at most $|\mathcal{U}_C|\sum_{i=1}^{m}\binom{|\mathcal{U}_C|}{i}$. Again, it should be noticed that the actual number of nodes is usually significantly smaller as isomorphic subgraphs are collapsed.

In case $m$-faults is extended to include system variables also, it can be shown that the variables can be ordered s.t. the number of nodes in the layers containing system nodes is exponential in $m$ but quadratic in the number of system variables if $m \leq max_{S \in \mathcal{U}_S} |\Pi_S|$ (see Figure 8).

Finally, as the single fault assumption no longer applies, the number of configurations consistent with $C_i = y$ and evidence $\bar{y}$ is given by:

$$\mathrm{Card}_B(C_i = y, \bar{y}) = \sum_{c_i \in \mathcal{C}_i} \sum_{l_i \in \mathcal{L}_i} \frac{v(c_i)_{\bar{y}}}{2^{\#l_i}},$$

where $\mathcal{C}_i$ is the set of nodes $c_i$ with an outgoing 1-arc, $\mathcal{L}_i$ is the set of distinct paths from the $c_i$ in question to the terminal node and $\#l_i$ is the number of arcs on such a path.

Having multiple faults also supports other frame-works like [de Kleer and Williams, 1987] and [Williams and Nayak, 1996]. For instance, in circuit diagnosis [de Kleer and Williams, 1987] uses a logical model of the system to be diagnosed and determines the next action based on expected Shannon entropy. To calculate the expected Shannon entropy they require the conditional probability of a set of failed components (termed a *candidate* in [de Kleer and Williams, 1987]) given some observation. As their framework does not yield an easy way to obtain this probability they use an approximation. In our framework the logical circuits can be represented as ROBDDs which makes the necessary probabilities easily available.

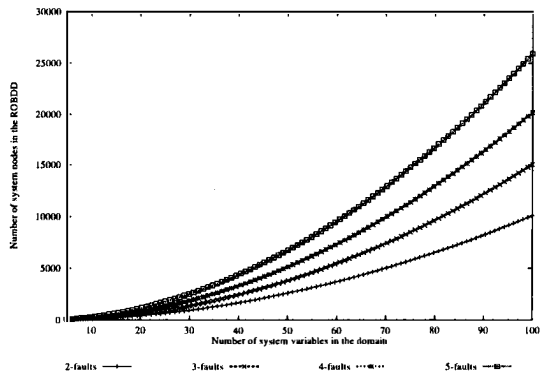

Figure 8: The number of nodes in the layers containing system nodes as a function of the number of system variables.

So far we have not established a practical upper bound on the size of ROBDDs with m faults, but all the examples we have worked with until now have been of a "small" size. Moreover, several heuristic methods have been devised for finding a good ordering of the variables (see e.g. [Malik et al., 1988] and [Fujita et al., 1988]).

## 9 CONCLUSION

When modelling the behavior of man-made machinery using Bayesian networks it frequently happens that a large part of the model is deterministic. In this paper we have reduced the task of belief updating in the deterministic part of such models to the task of calculating the number of configurations satisfying a Boolean function. In particular, we have exploited that a Boolean function can be represented by an ROBDD, and in this particular framework the number of satisfying configurations can be calculated in time linear in the size of the ROBDD.

The use of ROBDDs for belief updating was exemplified in the context of troubleshooting, which is particular well-suited as it was shown that the variables can be ordered s.t. the size of the ROBDD is quadratic in the size of the domain.

The performance of ROBDD propagation was compared with Shafer-Shenoy and Hugin propagation using randomly generated troubleshooting models. The results showed a substantial speed-up and it was argued that the single-fault assumption, underlying troubleshooting models, can be weakened without significantly affecting the performance of the algorithm in case the number of faults is "small".